\documentclass[conference,twoside]{IEEEtran}
\usepackage{cite}
\usepackage{color,colortbl}
\usepackage{graphicx}
\usepackage{amsmath}

\title{Extracting Tables from Documents using Conditional Generative Adversarial Networks and Genetic Algorithms}

\author{\IEEEauthorblockN{Nataliya Le Vine\IEEEauthorrefmark{1}, Matthew Zeigenfuse\IEEEauthorrefmark{1}, Mark Rowan\IEEEauthorrefmark{2}}

\IEEEauthorblockA{\IEEEauthorrefmark{1}Swiss Re, Digital and Smart Analytics, Armonk, New York, USA\\
}

\IEEEauthorblockA{\IEEEauthorrefmark{2}Swiss Re, Digital and Smart Analytics, Z\"urich, Switzerland\\
Email: \textit{\{nataliya\_levine,mark\_rowan\}@swissre.com, mdzeig@gmail.com}
}}

\definecolor{Gray}{gray}{0.9}

\begin{document}
\maketitle

\begin{abstract}
Extracting information from tables in documents presents a significant challenge in many industries and in academic research. Existing methods which take a bottom-up approach of integrating lines into cells and rows or columns neglect the available prior information relating to table structure. Our proposed method takes a top-down approach, first using a generative adversarial network to map a table image into a standardised `skeleton' table form denoting the approximate row and column borders without table content, then fitting renderings of candidate latent table structures to the skeleton structure using a distance measure optimised by a genetic algorithm.
\end{abstract}


\section{Introduction}
\label{introduction}
\subsection{Problem outline}
\IEEEPARstart{T}he Second Law of Thermodynamics states that the entropy of the Universe always increases: order flows into chaos and structure is lost. As anyone who works regularly with printed or digital documents can readily observe, this certainly also applies to the process of information storage and extraction in document form. Structured data is rendered as tables in documents, typically converting data into pixels of an image or characters on a page and thus stripping the data of meta-information relating to its structure. Information about hierarchical relationships of columns and rows is therefore lost.

Even in the digital era, common digital document exchange formats such as PDF do not explicitly represent any information about underlying data structures in tables. Instead, only individual characters and their locations on the page are encoded. Extracting and reimposing structure on data from tables in documents without reference to the original data structure from which they were created is a challenging problem in information extraction, particularly in document-based industries such as insurance and in the academic community where the primary method of knowledge transmission utilises printed or digital documents.

\subsection{Bottom-up vs top-down approaches}
\label{sec:bottom-up}
Existing approaches to the problem of extracting structured tabular information tend to be bottom-up. These approaches first identify pixel-level bounding boxes for table cells, exploiting regularities such as bounding lines and whitespace \cite{kasar2013learning,pinto2003table,gupta2019table}. The bounding boxes are subsequently aggregated into a table. Focusing on the graphical structure ignores available information relating to the higher-level structure of the table. As a consequence, bottom-up approaches have a tendency to be brittle and sensitive to specific layout quirks. In some cases, they can also depend heavily on correctly setting tuning parameters, such as the number of white pixels required to infer a cell boundary.

The approach by Klampfl et al. \cite{klampfl2014comparison} uses hierarchicaal agglomerative clustering to group information at the word layout level, and subsequently builds up a table from sets of words which are assumed to belong to a single column. As with the approaches described previously, this is a bottom-up approach which places the graphical representation of the layout as the highest priority.

A top-down approach, by contrast, assumes that a table in a document is merely an arbitrary graphical representation of an unseen latent data structure. Wang \cite{wang1996tabular} specifies a model in which the \textit{logical} and \textit{presentational} forms of a table are completely decoupled: given an underlying set of data and its hierarchical inter-relations (the logical table), the presentation of such a table in a document is totally arbitrary with regard to its formatting, style, rotation, column widths and row heights, etc.

For a specific graphical representation of a logical table in a document which we want to extract, we can attempt to fit arbitrary representations of candidate logical table structures onto the graphical representation seen in the document. By introducing a measure of distance between the representation of a candidate table structure and the representation seen on the page, a gradient can be followed towards an optimal solution. In this way, such an approach starts from the assumption that there is an underlying logical data structure which must be recovered, and decouples the search for the correct logical structure from the specifics of the graphical presentation, whilst using the graphical presentation to guide the search.

\subsection{Overview of proposed top-down approach}
\begin{figure}
    \includegraphics[width=\linewidth]{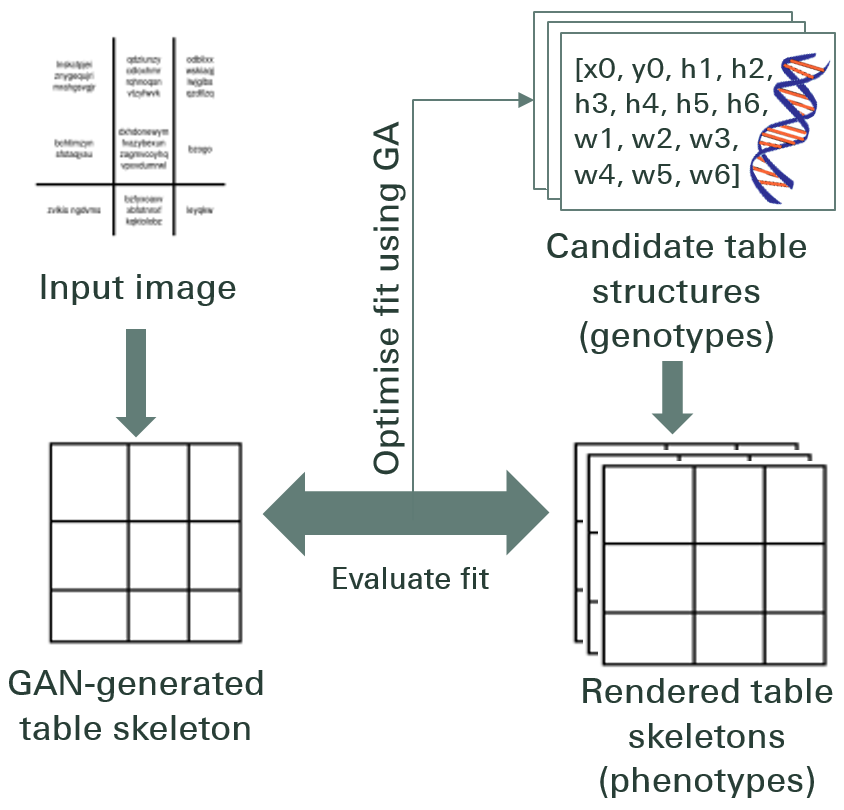}
    \caption{General schematic of the approach}
    \label{fig:skeleton}
\end{figure}

We propose a two-step generalisable top-down approach to extracting data from tables in documents (shown schematically in figure \ref{fig:skeleton}):
\paragraph{Step 1} translate input images containing tables into an abstracted standard ``skeleton'' form showing pixel outlines for approximate locations of table cells and boundaries and disregarding actual table content.
\paragraph{Step 2} optimise the fit of candidate latent data structures to the generated skeleton image using a measure of the distance between each candidate and the skeleton.

Once a good fit has been found, the data can be extracted from the table image and stored within the discovered structure using standard optical character recognition (OCR) techniques.

\subsection{cGAN and Genetic Algorithm}
A conditional generative adversarial network (cGAN) adversarially trains two neural networks: a \textit{generator} learns to generate realistic ``fake'' candidates to fool a \textit{discriminator} which learns to detect the artificially generated input out of a set of real input examples. cGAN has shown remarkable success at generating realistic looking images from input data, for example transforming from aerial photographs to street maps, from greyscale to colour, or filling in outlines with realistic detail \cite{isola2017image}. We treat Step 1 of our table extraction method as another form of image translation, by training a cGAN to translate images containing tables into a standardised skeletons denoting table cell borders. Our method for training the cGAN is outlined in section \ref{sec:scan_to_skeleton}.

Genetic algorithms (GA) use a population-based approach to sample the search space of possible solutions and to climb a gradient towards an optimum. The underlying representation of a candidate solution (dimensions and offsets of rows and columns of a table) is encoded as a vector of numbers denoted as the \textit{genotype}, which is decoded into a \textit{phenotype} (a rendered image of a table). Individuals within the population are then evaluated by a fitness function $f$ acting on the phenotypic representation. In our case, $f$ is provided by the \textit{discriminator} component of the cGAN which returns the distance between a rendered individual and a target image. Pairs of individuals with superior fitness (smaller distances to the target image) are probabilistically selected, and their genotypes \textit{recombined} to create new offspring solutions containing beneficial features of both parents -- for example combining the number of columns from one parent with the number of rows from the second parent. A genetic \textit{mutation operator} additionally maintains diversity within the population by randomly altering values by a rate $\mu$ to search the neighbouring solution space. Selection pressure thus acts at the phenotype level (comparison between rendered candidate solutions and the cGAN-generated target skeleton) to optimise the underlying genotype representation. The GA optimisation is outlined in section \ref{sec:ga}.

\subsection{Comparison to other approaches}
A comprehensive comparison versus state-of-the-art systems is out of the scope of this paper but nevertheless, to bring context to our approach we would like to comment on two representative commercial approaches: \textit{FlexiCapture} from Abbyy\footnote{http://help.abbyy.com/en-us/flexicapture/12/flexilayout\_studio/general\_tables} and \textit{CharGrid} from SAP \cite{katti2018chargrid}. \textit{FlexiCapture} is similar to other approaches we have seen, in that the user is required to manually define a hierarchical template for different ``classes'' of document (e.g. certain types of forms that the user regularly has to deal with), but the user is helped by some automatic estimation of table separators. This approach does not generalise well to previously unseen table layouts in new documents, unlike our approach which attempts to infer the latent data structure directly from the table image with no manual intervention.
\textit{CharGrid} \cite{katti2018chargrid} uses encoder-decoder convolutional neural networks to classify elements of a page and discover bounding boxes, including components of tables. This is in some ways similar to our approach for identifying table components visually in 2D space. In the case of tables, however, these elements must still be integrated together according to a bottom-up approach (see section \ref{sec:bottom-up}) based on character locations on the page. Whilst the \textit{CharGrid} representation for 2D text is powerful, it nevertheless does not make specific use of the prior assumptions relating to latent structure in table data which a top-down approach, such as we are proposing, can benefit from.

\section{Methods}
\label{method}
Table structure extraction is formulated as an image-to-genotype translation problem, whereby a table image is translated into a set of numbers (genotype) that fully characterise the table structure, i.e. column, row, and table positions and sizes. This is solved in two stages: 1) translating an input table image (scan) into a corresponding table cell border outline image (table `skeleton'), and 2) transforming the table `skeleton' into a latent table data structure, represented as a genetic algorithm genotype (figure \ref{fig:skeleton}). The first stage ignores any text present in a table image, keeps any table cell borders present on the image, and adds any missing cell separators which are implied by whitespace dividers in the input image. Then, the second stage selects the optimal latent data structure parameterisation based on an a-priori selected class of tables (e.g. whether cell merging is allowed across rows/columns, or not). Note that, while it can be relaxed, an underlying assumption is that there is no other text surrounding a table on an input image, and that all elements on an input image belong to one table.

First, an input table image is transformed into a table `skeleton' using a conditional Generative Adversarial Network (cGAN) \cite{isola2017image}. A conditional GAN consists of two adversarial mechanisms: 1) image generator G trained to transform an input image into an image similar to a target output image, and 2) image discriminator D trained to detect `fake' output images. Second, a table genotype describing the table geometry is optimised using a table generator, so that the corresponding table `skeleton' phenotype (table skeleton rendered as an image) is close to the table skeleton produced by the cGAN (figure \ref{fig:skeleton}).

\subsection{Table image to table skeleton}
\label{sec:scan_to_skeleton}
A table image is mapped into a corresponding table skeleton (all row and column dividers, even if not present on the table image; and text masked out) using cGAN.

\subsubsection{cGAN architecture}
First, to reduce cGAN model complexity (number of parameters) and to optimise image-processing time, a table scan is resized to a smaller image of 256x256. Then the resized table image is transformed into a table skeleton using the cGAN architecture from \cite{isola2017image}, as described in the paper.

A conditional GAN approximates a mapping from input image $x$ and random noise vector $z$ to output image $y$, so that $y=G(x,z)$. Generator $G$ is trained to produce outputs that cannot be distinguished from the target output images by a discriminator $D$ that is adversarially trained to detect `fake' images. Generator $G$ is an encoder-decoder network, so that an input image is passed through a series of progressively down-sampling layers until a bottleneck layer, where the process is reversed. To pass sufficient information to the decoding layers, a U-Net architecture with skip connections is used \cite{ronneberger2015u, isola2017image}, and a skip connection is added between each layer $i$ and $n-i$ via concatenation, where $n$ is the total number of layers, and $i$ is a layer number in the encoder \cite{isola2017image}. Further, following \cite{isola2017image}, random noise $z$ is provided only in the form of dropout applied on several layers of the generator at both train and test times.

The discriminator $D$ takes two images -- an input image and either a target or output image from the generator -- and assigns a probability that the second image is generated by a generator or not (i.e. the image is real). A convolutional PatchGAN architecture is used for the discriminator D \cite{li2016precomputed, isola2017image} that penalises output image structure at the scale of patches. In particular, the discriminator classifies whether each $N$ x $N$ patch in an image $y$ is real or fake ($N=70$), so that $D(x,y)$ equals $M$ x $M$ matrix containing probabilities for the patches in $y$ to be real ($M=30$), i.e. representative of a target image given input image $x$. Such a discriminator treats an image as a Markov random field, assuming independence between pixels separated by more than a patch diameter, and is understood as a form of texture loss.

\subsubsection{cGAN optimisation and inference}
The cGAN parameters are selected by optimising the following objective function with respect to generator $G$ and discriminator $D$:

\begin{equation}
	\label{eq:objective}
    \min_{G} \left[ \max_{D} L_{cGAN} (G,D) + \lambda L_{L1} (G) \right]
\end{equation}

Where $L_{L1}$ is $L1$ - distance between an image produced by a generator $G$ and an output image averaged over the training set, $\lambda$ is a constant weight ($\lambda=100$), and $L_{cGAN}$ expresses a cGAN objective to generate images that cannot be discriminated from real images:

\begin{equation}
    \begin{split}
    	\label{eq:lcgan}
        L_{cGAN} (G,D) &= E_{x,y} \left[\log D(x,y) \right]\\
                       &+ E_{x,y} \left[\log \left(1-D\left(x,G(x,z)\right)\right)\right]
    \end{split}
\end{equation}

Mixing of $L_{cGAN}$ and $L_1$ is found to be beneficial in previous studies, with $L_{cGAN}$ encouraging less blurring and $L_1$ suppressing artefacts \cite{pathak2016context, isola2017image}.

The cGAN is trained by alternating between optimising with respect to the discriminator $D$, and then with respect to the generator $G$ \cite{goodfellow2014generative}. Optimisation is performed on a set of 1,000 table scan/skeleton pairs (see section \ref{sec:traindata} for details of training set generation) using stochastic gradient descent applying the Adam solver \cite{kingma2014adam}, with learning rate 0.001 (decreasing to 0.0001 for a final set of epochs), and momentum parameters $\beta_1$=0.9, $\beta_2$=0.999. Following \cite{ulyanov2016instance} and \cite{isola2017image}, the generator $G$ is used with dropout and `'instance normalisation' (batch normalisation with a batch of 1) at inference time.

\subsection{Optimising genotype for a table skeleton}
\label{sec:skeleton_to_struct}
The derived table skeleton parameterises an objective function that defines table structure fitness to allow its optimisation by a Genetic Algorithm.

\subsubsection{Random table generator}
\label{sec:traindata}
A table structure is described as a set of the following numbers, defined as table genotype:
\begin{enumerate}
    \item table cardinality – number of rows $n$, and number of columns $m$;
    \item coordinates of the upper left corner of the table $\{x_0,y_0\}$;
    \item vector of row heights $\{h_1,\ldots,h_n\},h_i\geq0$;
    \item vector of column widths $\{w_1,\ldots,w_m\},w_i\geq$0.
\end{enumerate}
Number or rows $n$ and number of columns $m$ given a-priori express an expected maximum table cardinality. For a specific table, when a number of rows and columns is smaller than $n$ and $m$, respectively, the corresponding row heights $h_i$ and column widths $w_i$ are set to zero in the table genotype. The genotype represents only a simple 2-D table class in the case when merging of table cells is disabled. The table class can be (but not pursued in this study) further extended by introducing a cell merge indicator array $\{c_{ij}\}_{(i,j=1)}^{(n,m)}$ into the table genotype, where $c_{ij}$ is set to a unique number indicating what type of cell merging (if any) is needed for cell in $i$th row and $j$th column.

Each table genotype additionally parameterises an XHTML representation of the table which is rendered into an image using the imgkit Python library (table phenotype). Train and test sets for cGAN consist of table scan/skeleton image pairs, so that a table scan image requires adding random text into the table cells as well as choosing, at random, row and column separators to be visualised in the rendered image. The corresponding table skeleton excludes the text and shows all column and row separators. While optimising table genotype for a given table skeleton, requires only generating the corresponding table phenotypes (candidate table skeletons) that have all table separators visualised (even if not present on the original table scan), and text masked out on the image.

\subsubsection{Initial estimation}
\label{sec:initialguess}
Two types of optimisation starting points (initialisations) are tested: 1) random, and 2) cGAN projection-based. The random initialisation draws a set of candidate table genotypes -- each with a random number of rows and columns, random row heights and column widths, and random upper left corner coordinates -- using corresponding uniform distributions over feasible value space (uniform Monte-Carlo sampling). The cGAN-based initialisation projects a cGAN skeleton on x- and y-axis to estimate the corresponding table parameters using the xy-cut method \cite{nagy1984hierarchical}. The random initialisation requires a long optimisation time, and often results in a local optimum, while the cGAN projection-based optimisation often requires only several optimisation cycles to reach an optimum (see the Results section). For this reason, only the cGAN  projection-based initialisation is used in the following sections.

\subsubsection{Optimisation with genetic algorithm}
\label{sec:ga}
The initial table structure guess is further optimised using Genetic Algorithm (GA). GA uses reproduction, crossover, and mutation to evolve a population of tables (population size = 50) from epoch $n-1$ to epoch $n$, so that table structure fitness is improved following the gradient of an objective function. Each table has its own genotype and phenotype as defined in section \ref{sec:traindata}, and the algorithm selects candidate tables based on the table fitness defined in section \ref{sec:ga_objective}. GA requires specifying a number of parameters (e.g. mutation rate, survival rate) selected experimentally in the study. The algorithm converges when table fitness does not improve more than 1\% over three consecutive epochs.

In the algorithm, reproduction carries the best table structure over to the next epoch with no mutations (elitism), as well as 70\% (values between 30\% and 90\% were tested) of other table structures with mutation. Further, the offspring mutation modifies table upper left corner coordinates, individual row heights, and column widths with a probability 0.1 (values between 0.05 and 0.3 were tested) for each entry in the genotype. The offspring mutation also modifies table structure (adding, merging, removing column/row) with probability of 0.1 per table dimension (columns or rows), so that the three structural operations are equally likely (probability of 0.03). Lastly, crossover is based on two parents, so that the upper left table x-coordinate and columns are inherited from the first parent, while the upper left table y-coordinate and rows are inherited from the second parent.

\subsubsection{Objective functions for genotype optimisation}
\label{sec:ga_objective}
Table fitness is defined linearly proportional to an objective function value for the table, so that a fitter table has a better objective function value. Several objective functions are tested to measure fit between a cGAN generated table skeleton and the rendering of a candidate table phenotype $u$ representing an optimised table genotype:
\begin{enumerate}
	\item Maximising probability of a candidate table phenotype $u$ to be true according to the cGAN discriminator:
	\begin{equation}
		\label{eq:max_w}
        \max_{u}\left[⁡\log D(x,u)\right]
	\end{equation}

	\item Minimising the $L_1$ distance between cGAN generated image and a candidate image at pixel level:
	\begin{equation}
		\label{eq:min_l1}
		\min_{u}|G(x,z)-u|_{L1}
	\end{equation}

    \item Maximising a weighted difference between (\ref{eq:max_w}) and (\ref{eq:min_l1}) similar to (\ref{eq:objective}) as:
	\begin{equation}
		\label{eq:weighted}
		\max_{u}\left[log D(x,u)-\lambda|G(x,z)-u|_{L1}\right]
	\end{equation}

	\item Minimising fraction of non-overlapping non-white pixels between cGAN and the candidate image, calculated as:
	\begin{equation}
        \label{eq:min_overlap}
        \min_{u} \frac{|G(x,z)-u|_{L1}} {|1-u|_{L1} \cdot |1-G(x,z)|_{L1}}
    \end{equation} where pixels of the output image from cGAN $G(x,z)$ and a candidate table phenotype are scaled to values between 0 and 1, with 0 corresponding to a black pixel, and 1 corresponding to a white pixel. The objective function \ref{eq:min_overlap} varies between 0 and 1, so that the best match corresponds to 0 and the worst guess corresponds to 1.
\end{enumerate}

Maximising probability of a candidate table to be true (\ref{eq:max_w}) is expected to help with introducing table configurations that may differ from the cGAN generated table skeleton. However, optimising (\ref{eq:max_w}) leads to development of unexpected table configurations (often with larger than expected number of columns and rows), characterised with large changes in a table candidate, but minor changes in the objective function values.

Minimising the $L_1$ distance between cGAN output image and a candidate image (\ref{eq:min_l1}) favors smaller number of rows and columns than optimal configuration requires. This is due to the objective function (\ref{eq:min_l1}) specifics: minimising the number of column/rows (number of non-white pixels) in a candidate table decreases the number of non-overlapping non-white pixels (column/row borders) between cGAN skeleton and candidate skeleton (as there are fewer non-matching non-white pixels), and hence this improves the objective function (\ref{eq:min_l1}). Consequently, the objective function is drawn to its local minima when a candidate table has minimum possible number or rows and columns. Meanwhile, finding the function global minima is a `needle-in-a-hay-stack' problem as the objective function is discontinuous (hence no gradient to follow in optimisation) when table skeletons in the cGAN and candidate image match exactly. Moreover, the objective function (\ref{eq:min_l1}) has a low sensitivity to changes in table structure as adding or removing a column/row only affects a small fraction of image pixels. This leads to a low differentiability between candidate table skeletons in the GA algorithm, when structurally different tables have similar fitness values (proportional to objective function values). 

Objective function (\ref{eq:weighted}) combines objective functions (\ref{eq:max_w}) and (\ref{eq:min_l1}) attempting to overcome their individual limitations -- in particular, cardinality expansion in (\ref{eq:max_w}) and cardinality shrinkage in (\ref{eq:min_l1}). However, experimenting with a range of weights $\lambda$=1,10,100 in (\ref{eq:weighted}) did not provide a reasonable objective function behavior as corresponding candidate table configurations were not found to gradually improve, but rather they developed unexpectedly.

Testing the three objective functions (\ref{eq:max_w})--(\ref{eq:weighted}) motivated developing a new objective function (\ref{eq:min_overlap}). The function is designed to mitigate the low sensitivity problem in function (\ref{eq:min_l1}), and to penalise incorrectly adding or removing rows/columns to the table. Its drawback is that (\ref{eq:min_overlap}) does not favor table structures that differ from a generated cGAN skeleton structure, so that the optimum table identification depends on a cGAN generation quality. This has the effect of reducing the role of the genetic algorithm to fine-tuning the output of the cGAN, rather than fully exploring the search space of potential candidates. Results in the following sections use this objective function.

\section{Experiments}
\label{sec:results}
Experiments were conducted for table structure estimation using: 1) GA for one type of table configuration (parameters defining font, number of rows and columns, etc.); 2) initial cGAN estimate only (no GA optimisation) for the same type of table configuration; 3) initial cGAN estimate for multiple table configurations. The latter experiment was performed due to the model's sensitivity to some specific table parameters such as intra-column/row padding and text spacing.

\begin{figure}
    \center
    \includegraphics[width=\linewidth]{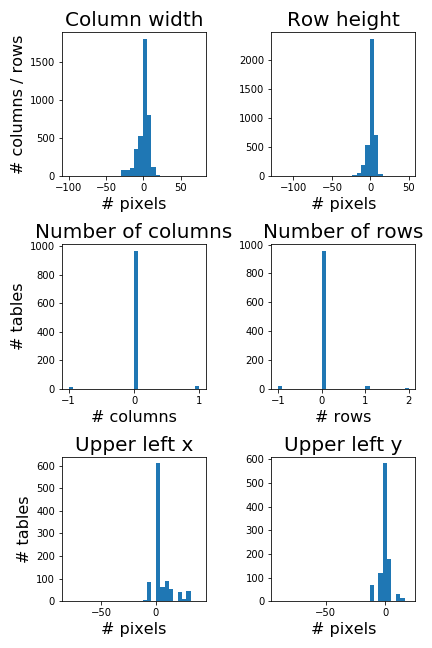}
    \caption{Histograms of the evaluation metrics for GA optimised table structures for 1,000 test table scans with the base configuration.}
    \label{fig:histograms}
\end{figure}

\begin{figure}[t]
    \center
    \includegraphics[width=0.8\linewidth]{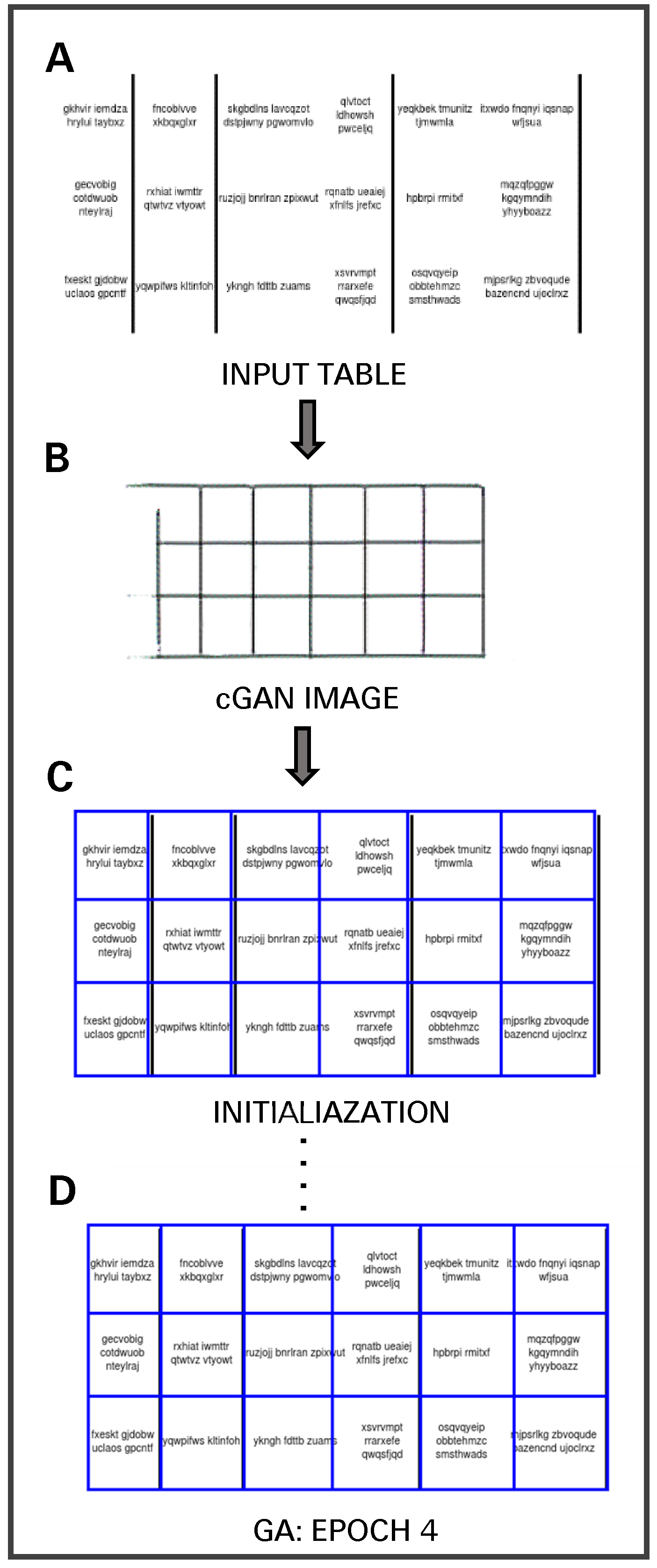}
    \caption{Table structure estimation example: input table scan is transformed into a table `skeleton' draft by cGAN; then GA is used to optimise table genotype for the corresponding table phenotype to fit the cGAN `skeleton'. Initialisation and Epoch 4 tables are overlays of the table scan (black) and best available table structure estimate (blue lines).}
    \label{fig:results}
\end{figure}

\subsection{Evaluation metrics and performance}
The proposed method is applied to images measuring 595x842 pixels (corresponding to A4 paper size at 72 ppi) generated by the random table generator described in section \ref{sec:traindata} with parameters given in table \ref{tab:tableconfigs} in the `base configuration' row. Conditional GAN is trained using 1,000 such table images, and a further 1,000 images are generated for performance evaluation. Genetic algorithm with objective function (\ref{eq:min_overlap}) is used to optimise an initial table structure estimate until the change in objective function value over three consecutive epochs is less than 1\%.

The estimated table structures are evaluated by comparing: 1) row and column number, 2) upper left corner position, 3) row heights and column widths, and the results are shown in table \ref{tab:tableconfigs} (given in the `Single type model' column, `GA' sub-column) and figure \ref{fig:histograms}. While the model generally provides skillful table structure estimates, in less than 5\% of cases there is a column missing or unnecessarily added, or a row missing or 1-2 rows added. Moreover, one of the observed peculiarities is that sometimes 1-2 of the furthest left or right characters in a cell get cut off by the table column divider (see figure \ref{fig:results}), which might lead to errors on a potential following OCR step. Very narrow whitespace buffers between table dividers and cell contents in the generated tables further enhance this effect.

The initial table structure estimation described in section \ref{sec:initialguess} usually provides a good approximation of the final table structure found by the GA. Furthermore, in most cases, the GA only slightly moves column and row dividers in the table to align the best with a corresponding GAN skeleton as quantified by the objective function (\ref{eq:min_overlap}) (compare Initialisation (figure \ref{fig:results}C) and Epoch 4 (figure \ref{fig:results}D)). Hence, we additionally evaluated only the table structure initial guess, i.e. with no further GA optimisation, using the same set of test table scans as above, and the results are shown in Table 2 (column `Single type model', `Initial guess' sub-column). The resulting metrics are comparable to those from the GA optimisation, with minor differences explained by small variations in the generated cGAN skeletons due to the randomness in the dropout applied, as well as minor row and column divider position adjustments by the GA (as discussed above). As an initial estimate is more time efficient (no generating table population, applying GA operations to them, and evolving to the next epoch), we turned off the GA and used only the initial cGAN projection in the further experiments.

\subsection{Testing on more varied table specifications}
\begin{table*}[t]
    \centering
    \caption{Table configurations for sensitivity tests. ``$\cdot$'' refers to values the same as the `Base' configuration. Shaded rows denote table configurations to which the cGAN was found to be sensitive.}
    \label{tab:tableconfigs}
    \scriptsize
    \begin{tabular}{|p{1.7cm}|p{0.5cm}p{0.5cm}p{0.8cm}p{0.8cm}p{0.8cm}p{1cm}p{0.8cm}p{0.8cm}p{1.3cm}p{0.8cm}p{1.2cm}|}
        \hline
        \textbf{Configuration} & \textbf{Rows} & \textbf{Cols} & \textbf{x-offset, ptx} & \textbf{y-offset, ptx} & \textbf{Row height, px} & \textbf{Column width, px} & \textbf{Word length, chars} & \textbf{Words per cell} & \textbf{Font family} & \textbf{Font size, px} & \textbf{Alignment}\\
        \hline
        Base            & 2--6 & 2--6 & 0--70 & 0--70 & 40--90 & 70--100 & 5--9 & 2--4 & Arial & 10 & center \\
        Font 1          & $\cdot$ & $\cdot$ & $\cdot$ & $\cdot$ & $\cdot$ & $\cdot$ & $\cdot$ & $\cdot$ & New Roman & $\cdot$ & $\cdot$ \\
        Font 2          & $\cdot$ & $\cdot$ & $\cdot$ & $\cdot$ & $\cdot$ & $\cdot$ & $\cdot$ & $\cdot$ & Courier & $\cdot$ & $\cdot$ \\
        Larger font 1   & $\cdot$ & $\cdot$ & $\cdot$ & $\cdot$ & $\cdot$ & $\cdot$ & $\cdot$ & $\cdot$ & $\cdot$ & 14 & $\cdot$ \\
        \rowcolor{Gray}
        Larger font 2   & $\cdot$ & $\cdot$ & $\cdot$ & $\cdot$ & $\cdot$ & $\cdot$ & $\cdot$ & $\cdot$ & $\cdot$ & 18 & $\cdot$ \\
        \rowcolor{Gray}
        Smaller font    & $\cdot$ & $\cdot$ & $\cdot$ & $\cdot$ & $\cdot$ & $\cdot$ & $\cdot$ & $\cdot$ & $\cdot$ & 6 & $\cdot$ \\
        Skinny long cells & $\cdot$ & $\cdot$ & $\cdot$ & $\cdot$ & 20 & 120--200 & $\cdot$ & 3--7 & $\cdot$ & $\cdot$ & $\cdot$ \\
        \rowcolor{Gray}
        Short cells     & 4--10 & 4--10 & $\cdot$ & $\cdot$ & 20 & 40--60 & 1--4 & 1 & $\cdot$ & $\cdot$ & $\cdot$ \\
        Align left      & $\cdot$ & $\cdot$ & $\cdot$ & $\cdot$ & $\cdot$ & $\cdot$ & $\cdot$ & $\cdot$ & $\cdot$ & $\cdot$ & left \\
        Align right     & $\cdot$ & $\cdot$ & $\cdot$ & $\cdot$ & $\cdot$ & $\cdot$ & $\cdot$ & $\cdot$ & $\cdot$ & $\cdot$ & right \\
        \hline
    \end{tabular}
\end{table*}

\begin{table*}[t]
    \centering
    \caption{Metrics for two different model types (single-type and multi-type tables), two different optimisation stages (initial guess, GA), and different model table configurations (base, smaller font, larger font 2, and short cells configurations). Errors are for absolute differences, except for number of rows/columns, and are shown as mean values with standard deviation in parentheses. Errors in number of rows/columns are calculated as true value less predicted value.}
    \label{tab:results}
    \scriptsize
    \begin{tabular}{|l|lllll|llll|}
        \hline
         & \multicolumn{5}{c}{\textbf{Single-type model}} & \multicolumn{4}{c|}{\textbf{Multi-type model}} \\
        \textbf{Metric} & \textbf{GA:base} & \textbf{Base} & \textbf{Smaller font} & \textbf{Larger font} & \textbf{Short cells} & \textbf{Base} & \textbf{Smaller font} & \textbf{Larger font 2} & \textbf{Short cells} \\
        \hline
        \% correct row count & 95.5 & 97.1 & 32.7 & 19.4 & 4.5 & 72.5 & 93.6 & 83.5 & 98.6 \\
        \% correct column count & 96.7 & 96.8 & 22.1 & 55.3 & 3.6 & 78.3 & 92.3 & 77.3 & 85.0 \\
        Error in row number & 0.1 (0.3) & 0.0 (0.2) & 1.1 (1.1) & -1.8 (1.4) & 3.0 (1.7) & 0.2 (0.6) & 0.1 (0.3) & -0.2 (0.4) & 0.0 (0.1) \\
        Error in column number & 0.0 (0.2) & 0.0 (0.2) & 1.3 (1.0) & -0.7 (1.0) & 3.3 (1.9) & 0.1 (0.5) & 0.0 (0.3) & -0.1 (0.5) & 0.2 (0.4) \\
        Error in x0, px & 5.1 (8.7) & 5.1 (9.2) & 6.5 (14.8) & 4.1 (5.9) & 4.5 (12.2) & 4.7 (14.7) & 2.7 (5.0) & 8.0 (19.6) & 2.0 (3.6) \\
        Error in y0, px & 3.2 (6.6) & 3.4 (6.1) & 9.1 (11.8) & 6.5 (8.2) & 8.4 (9.8) & 7.2 (11.2) & 3.2 (3.8) & 6.5 (7.2) & 2.9 (4.0) \\
        Error in col. width, px & 5.5 (8.0) & 5.4 (7.9) & 27.7 (36.3) & 50.7 (31.0) & 45.3 (58.2) & 10.2 (18.5) & 5.3 (7.1) & 39.7 (31.8) & 5.1 (8.8) \\
        Error in row height, px & 3.4 (7.7) & 3.1 (7.7) & 24.6 (36.0) & 59.4 (44.8) & 16.7 (21.6) & 10.4 (21.7) & 3.4 (5.8) & 17.1 (27.9) & 0.5 (1.5) \\
        \hline
    \end{tabular}
\end{table*}

\begin{figure}
    \includegraphics[width=\linewidth]{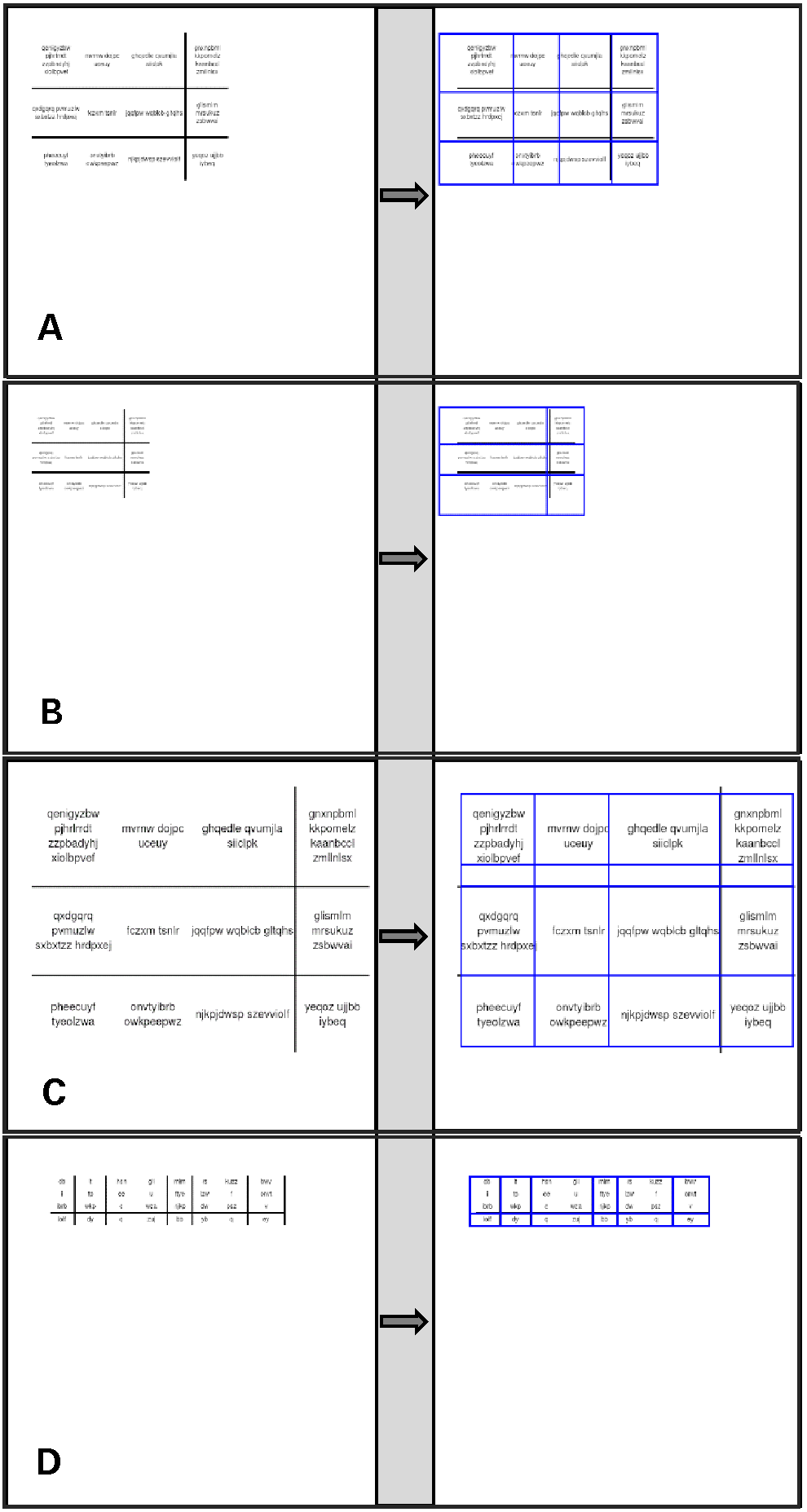}
    \caption{Four table configurations and typical errors identified by the sensitivity analysis: A. `base', B. `smaller font', C. `larger font 2', and D. `short cells'. Input table scans are in the top panel and blue output table structures overlaid over the table inputs are in the bottom panel (all images are cropped to squares).}
    \label{fig:variations}
\end{figure}

The model developed in the previous section is based on the `base' table configuration described in table \ref{tab:tableconfigs}, and hence the model is not guaranteed to extrapolate to other configurations. To assess the impact of using only a single table configuration, nine perturbations are tested: smaller and larger font sizes, different font families, text alignment in a cell, and variations in cell width and height (table \ref{tab:tableconfigs}). First, a manual sensitivity analysis based on five tables of each kind identified three table configurations that the model incorrectly estimates: `smaller font', `larger font 2', and `short cells' (figure \ref{fig:variations}). For illustration purposes, figures \ref{fig:variations}A-C show these configurations with the same number of rows and columns as well as identical textual content, but different font size, row/column padding, and word spacing. Due to the configuration specifics, it was not possible to match this for the `short cells' configuration shown in figure \ref{fig:variations}D.  Table \ref{tab:results} shows model performance on the three sensitive table configurations for 1,000 tables of each kind (in the `Single type model' column). The two main issues observed here were that the model tends to miss rows and columns for the `small font' and `short cell' configurations (figures \ref{fig:variations}B\&D), and inserts unnecessary columns and rows for the `larger font 2' configuration (figure \ref{fig:variations}C.) The latter has a low impact on a further OCR step, as OCR can help in detecting and removing empty cells.

Second, another cGAN model is trained on 4,000 samples with equal number of samples from each of `base', `smaller font', `larger font 2', and `short cells' table configurations; and the derived model is assessed on the four table configurations using an independent test set from the previous testing. Model performance improves significantly on the three added table configurations, mainly due to a better identification of rows and columns; but deteriorates for the `base' configuration, where the number of missing rows and columns increases (`Multi-type model' column in table \ref{tab:results}). This might be because the cGAN model parameters may be located at a local optima, or perhaps due to difficulty in visually discerning columns and rows when row/column padding, word spacing, and overall image texture (pattern) change.

\section{Conclusions}
\label{sec:conclusions}
The study addresses the problem of table structure estimation from scanned table images. The problem can have several valid solutions, as table row and column divider positions can be slightly perturbed and still accurately separate cell contents, due to padding around the dividers between rows and columns. For simplicity, the study focuses on a class of tables that does not contain merged cells; a further extension to a wider class of tables is theoretically possible, but not tested here.

The problem is re-formulated as an image-to-genotype translation, and is solved in two stages: 1) translating an input table image into a corresponding table `skeleton' using cGAN, and 2) fitting a latent table structure (table genotype) to the derived table `skeleton' using Genetic Algorithm. Conditional GAN output allows parameterising an objective function measuring a genotype's fitness, which is optimised by GA. The derived solution has the following properties:
\begin{itemize}
    \item Due to the close resemblance between a cGAN-generated table skeleton and the target table structure, an initial table genotype estimate based on the cGAN skeleton provides skillful table structure estimates.
    \item A further optimisation of the table structure with GA adjusts row and column divider positions (by several pixels), but does not provide significant improvements over the initial estimate due to the choice of objective function (\ref{eq:min_overlap}), which causes the GA to optimise along a gradient which leads towards the output of the cGAN. The GA is therefore mainly concerned with fine-tuning the match between latent table structure estimates (table genotypes) and the generated table skeleton image, rather than fully exploring the potential solution space of genotypes beyond the initial cGAN-generated table estimate. Potentially, a simpler gradient-ascent algorithm could therefore be used in place of the GA.
    \item However, the improvements from the GA fine-tuning might still be of importance for a potential next OCR step, especially if there is little padding between cell content and column dividers.
    \item The cGAN model is sensitive to table and cell text specifics, in particular large variations in font size, row/column padding, and word spacing.
    \item A model trained on a wider range of table configurations performs well overall, but does not provide an equally good performance for all configurations considered.
\end{itemize}
Improving the solution further requires an extensive optimal cGAN parameter search with random search initialisations to improve the current locally optimal solution, needs a more complex computer vision model, or needs a computer vision / NLP model hybrid solution. Further, to be used in practice, the solution would need to be combined with an algorithm to detect the table area for tables in documents surrounded by text, as well as combined with an OCR algorithm to extract text into the data structure. Lastly, the described cGAN application can be further extended to delineating more general shape patterns, such as shape compositions, graphs, and embedded images.

\section{Acknowledgements}
We thank Dr. Gianluca Antonini and Dr. Claus Horn for their reviews and helpful comments to improve the manuscript.

\newpage

\section{References}
\bibliographystyle{plain}
\bibliography{refs}

\begin{thebibliography}{10}

\bibitem{goodfellow2014generative}
Ian Goodfellow, Jean Pouget-Abadie, Mehdi Mirza, Bing Xu, David Warde-Farley,
  Sherjil Ozair, Aaron Courville, and Yoshua Bengio.
\newblock Generative adversarial nets.
\newblock In {\em Advances in neural information processing systems}, pages
  2672--2680, 2014.

\bibitem{gupta2019table}
Anand Gupta, Devendra Tiwari, Tarasha Khurana, and Sagorika Das.
\newblock Table detection and metadata extraction in document images.
\newblock In {\em Smart Innovations in Communication and Computational
  Sciences}, pages 361--372. Springer, 2019.

\bibitem{isola2017image}
Phillip Isola, Jun-Yan Zhu, Tinghui Zhou, and Alexei~A Efros.
\newblock Image-to-image translation with conditional adversarial networks.
\newblock In {\em 2017 IEEE Conference on Computer Vision and Pattern
  Recognition (CVPR)}, pages 5967--5976. IEEE, 2017.

\bibitem{kasar2013learning}
Thotreingam Kasar, Philippine Barlas, Sebastien Adam, Cl{\'e}ment Chatelain,
  and Thierry Paquet.
\newblock Learning to detect tables in scanned document images using line
  information.
\newblock In {\em ICDAR}, pages 1185--1189, 2013.

\bibitem{katti2018chargrid}
Anoop~R Katti, Christian Reisswig, Cordula Guder, Sebastian Brarda, Steffen
  Bickel, Johannes H{\"o}hne, and Jean~Baptiste Faddoul.
\newblock Chargrid: Towards understanding 2d documents.
\newblock In {\em Proceedings of the 2018 Conference on Empirical Methods in
  Natural Language Processing}, pages 4459--4469, 2018.

\bibitem{kingma2014adam}
Diederik~P Kingma and Jimmy Ba.
\newblock Adam: A method for stochastic optimization.
\newblock {\em arXiv preprint arXiv:1412.6980}, 2014.

\bibitem{klampfl2014comparison}
Stefan Klampfl, Kris Jack, and Roman Kern.
\newblock A comparison of two unsupervised table recognition methods from
  digital scientific articles.
\newblock {\em D-Lib Magazine}, 2014.

\bibitem{li2016precomputed}
Chuan Li and Michael Wand.
\newblock Precomputed real-time texture synthesis with markovian generative
  adversarial networks.
\newblock In {\em European Conference on Computer Vision}, pages 702--716.
  Springer, 2016.

\bibitem{nagy1984hierarchical}
George Nagy and Sharad Seth.
\newblock Hierarchical representation of optically scanned documents.
\newblock In {\em Proc. 7th Int. Conf. Pattern Recognition}, pages 347--349,
  1984.

\bibitem{pathak2016context}
Deepak Pathak, Philipp Krahenbuhl, Jeff Donahue, Trevor Darrell, and Alexei~A
  Efros.
\newblock Context encoders: Feature learning by inpainting.
\newblock In {\em Proceedings of the IEEE Conference on Computer Vision and
  Pattern Recognition}, pages 2536--2544, 2016.

\bibitem{pinto2003table}
David Pinto, Andrew McCallum, Xing Wei, and W~Bruce Croft.
\newblock Table extraction using conditional random fields.
\newblock In {\em Proceedings of the 26th annual international ACM SIGIR
  conference on Research and development in informaion retrieval}, pages
  235--242. ACM, 2003.

\bibitem{ronneberger2015u}
Olaf Ronneberger, Philipp Fischer, and Thomas Brox.
\newblock U-net: Convolutional networks for biomedical image segmentation.
\newblock In {\em International Conference on Medical image computing and
  computer-assisted intervention}, pages 234--241. Springer, 2015.

\bibitem{ulyanov2016instance}
Dmitry Ulyanov, Andrea Vedaldi, and Victor Lempitsky.
\newblock Instance normalization: The missing ingredient for fast stylization.
\newblock {\em arXiv preprint arXiv:1607.08022}, 2016.

\bibitem{wang1996tabular}
Xinxin Wang.
\newblock {\em Tabular Abstraction, Editing, and Formatting}.
\newblock PhD thesis, University of Waterloo, 1996.

\end{thebibliography}
\end{document}